\documentclass[11pt,letterpaper]{article}
\usepackage{naaclhlt2016}
\usepackage{times}
\usepackage{latexsym}
\usepackage{graphicx}
\naaclfinalcopy 
\def\naaclpaperid{***} 

\title{Senti17 at SemEval-2017 Task 4: Ten Convolutional Neural Network Voters for Tweet Polarity Classification}

\author{Hussam Hamdan\\Labex Observatoire de la vie littéraire (OBVIL)\\
 Laboratoire d'Informatique de Paris 6 (LIP6), Pierre and Marie Curie University, UMR 7606\\
 4 place Jussieu, 75005, Paris, France\\
 {Hussam.Hamdan@lip6.fr}}

\date{}

\begin{document}

\maketitle

\begin{abstract}
This paper presents Senti17 system which uses ten convolutional neural networks (Con- vNet) to assign a sentiment label to a tweet. The network consists of a convolutional layer followed by a fully-connected layer and a Soft- max on top. Ten instances of this network are initialized with the same word embeddings  as inputs but with different initializations for the network weights. We combine the results of all instances by selecting the sentiment label given by the majority of the ten voters. This system is ranked fourth in SemEval-2017 Task4 over 38 systems with 67.4\% average recall.
\end{abstract}

\section{Introduction}

Polarity classification is the basic task of sentiment analysis in which the polarity of a given text should be classified into three categories: positive, negative or neutral. In Twitter where the tweet is short and written in informal language, this task needs more attention. SemEval has proposed the task of Message Polarity Classification in Twitter since 2013, the objective is to classify a tweet into one of the three polarity labels \cite{rosenthal_semeval-2017_2017}. 

We can remark that in 2013, 2014 and 2015 most best systems were based on a rich feature extraction process with a traditional classifier such as SVM \cite{mohammad_nrccanada:_2013} or  Logistic regression \cite{hamdan_lsislif:_2015}.
 In 2014, \newcite{kim_convolutional_2014} proposed to use  one convolutional neural network for sentence classification, he fixed the size of the input sentence and concatenated  its word embeddings for representing the sentence, this architecture has been exploited in many later works. 
 \newcite{severyn_unitn:_2015} adapted  the convolutional network proposed by  \newcite{kim_convolutional_2014} for sentiment analysis in Twitter, their system was ranked second in SemEval-2015 while the first system \cite{hagen_webis:_2015} combined four systems based on feature extraction  and the third ranked system used logistic regression with different groups of features  \cite{hamdan_lsislif:_2015}. 
 
 In 2016, we remark that the number of participations which use feature extraction systems were degraded, and the first four systems used Deep Learning, the majority used a convolutional network  except the fourth one \cite{amir_inesc-id_2016}. Despite of that, using Deep Learning for sentiment analysis in Twitter has not yet shown a big improvement in comparison to feature extraction, the fifth and sixth systems \cite{hamdan_sentisys_2016} in 2016  which were built upon feature extraction process were  only (3 and  3.5\% respectively) less than the first system. But We think that Deep Learning is  a promising direction in sentiment analysis. Therefore, we proposed to use convolutional networks for Twitter polarity classification.
 
Our proposed system consists of a convolutional  layer followed by fully connected layer and a softmax on top. This is inspired by \newcite{kim_convolutional_2014}, we just added a fully connected layer. This architecture gives a good performance but it could be improved. Regarding the best system in 2016 \cite{deriu_swisscheese_2016}, it uses different word embeddings for initialisation then it combines the predictions of different nets using a meta-classifier, Word2vec and Glove   have been used to vary the tweet representation. 

 In our work, we propose to vary the neural network weights instead of tweet representation which can get the same effect of varying the word embeddings, therefore we vary the initial weights of the network to produce ten different nets, a voting system over the these ten voters will decide the sentiment label for a tweet.  

The remaining of this paper is organized as follows: Section 2 describes the system architecture, Section 3 presents our experiments and results and  Section 4 is devoted for the conclusion.

\section{System Architecture}
The architecture of our convolutional neural net- work for sentiment classification is shown on Fig. 1.
Our network is composed of a single convolutional layer followed by a non-linearity, max pooling, Dropout, fully connected layer  and a soft-max classification layer. Here we describe this architecture: 

\subsection{Tweet Representation}
 We first tokenize each tweet to get all terms using HappyTokenizer\footnote{http://sentiment.christopherpotts.net/tokenizing.html} which captures the words, emoticons and punctuations. We also replace each web link by the term \textit{url} and each user name by \textit{uuser}. Then, we used  Structured Skip-Gram embeddings (SSG) \cite{ling_two/too_2015}  which was compiled by \cite{amir_inesc-id_2016} using 52 million tweets. 
 
Each term in the tweet is  replaced by its SSG embedding which is a vector of  \textit{d}  dimensions, all term vectors are concatenated to form the input matrix where the number of rows is \textit{d}  and the number of columns is set to be maxl: the max tweet length in the training dataset. This 2-dim matrix is the input layer for the neural network.

\subsection{Convolutional Layers}
We connect the input matrix with different convolutional layers, each one  applies a convolution operation between the input matrix and a filter of size \textit{m} x \textit{d}. This is an element-wise operation which creates \textit{f} vectors of  \textit{maxl-m+1} dimension where  \textit{f} is the number of filters or feature maps. 

This layer is supposed to capture the common patterns among the training tweets which have the same filter size but occur at any position of the tweet. To capture the common patterns which have different sizes we have to use more than one  layer therefore we defined 8 different layers connected to the input matrix with different filter sizes but the same number of feature maps. 
\begin{figure*}
  \includegraphics[width=\textwidth,height=6cm]{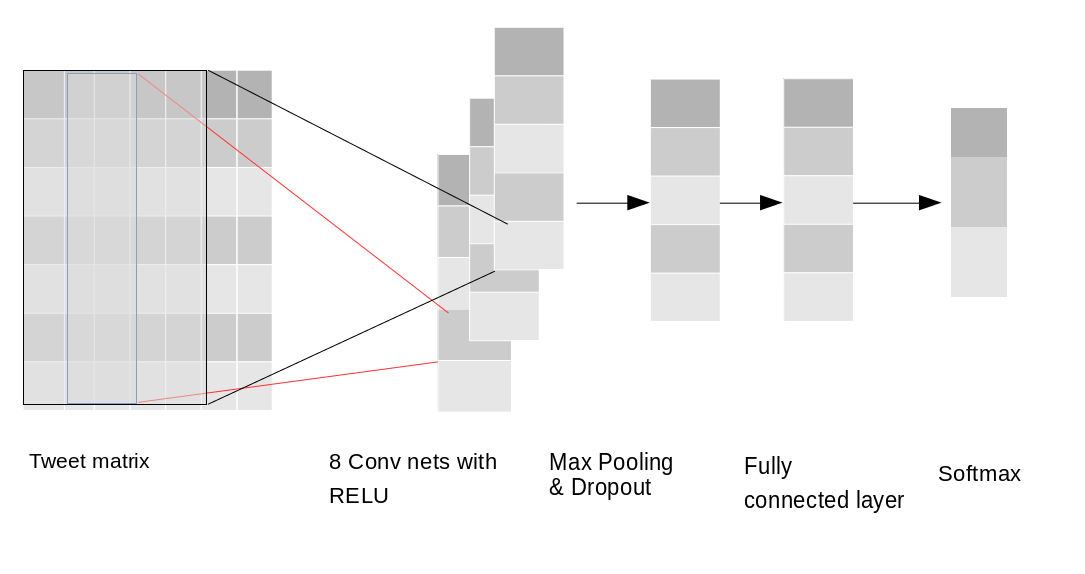}
  \caption{Network architecture.}
\end{figure*}

\subsection{Activation Layer}
Each convolutional layer is typically followed by a non-linear activation function, RELU (Rectified Linear Unit ) layer will apply an element-wise operation to swap the negative numbers  to 0. The output of a ReLU layer is the same size as the input, just with all the negative values removed. It speeds up the training and is supposed to produce more accurate results.
\subsection{Max-Pooling Layer}
This layer reduces the size of the output of activation layer, for each vector it selects the max value. Different variation of pooling layer can be used: average or k-max pooling. 
\subsection{Dropout Layer}
Dropout is used after the max pooling to regularize the ConvNet and  prevent  overfitting. It assumes that we can still obtain a reasonable classification even when some of the neurones are dropped. Dropout consists in randomly setting a fraction \textit{p} of input units to 0 at each update during training time. 
\subsection{Fully Conected Layer}
We concatenate the results of all pooling layers after applying Dropout,  these units are connected to a fully connected layer.  This layer performs a matrix multiplication between its weights and the input units. A  RELU non-linarity is applied on  the results of this layer.
\subsection{Softmax Layer}

The output of the fully connected layer is passed to a Softmax layer. It computes the probability distribution over the labels in order to decide the most probable label for a tweet.

\section{Experiments and Results}
For training the network, we used about 30000 English tweets provided by SemEval organisers and the test set of 2016 which contains 12000 tweets as development set. The test set of 2017 is used to evaluate the system in SemEval-2017 competition. For implementing our system we used python and Keras\footnote{https://keras.io}. 

We set the network parameters as follows: SSG embbeding size \textit{d} is chosen to be 200, the tweet max legnth \textit{maxl} is 99. For convolutional layers, we set the  number of   feature maps  \textit{f} to 50  and used 8 filter sizes (1,2,3,4,5,2,3,4). The \textit{p} value of Dropout layer is set to 0.3. We  used Nadam optimizer \cite{dozat_incorporating_2015}   to update the weights of the network and back-propogation algorithm to compute the gradients. The batch size is set to be 50 and the training data is shuffled after each iteration. 

We create ten instances of this  network,  we randomly initialize them using the uniform distribution, we repeat the random initialization for each instance 100 times, then we pick the networks which gives the highest average recall score as it is considered  the official measure for system ranking. If the top network of each instance gives more than 95\% of its results identical to another chosen network, we choose the next top networks to make sure that the ten networks are enough different.

Thus, we have ten classifiers, we count the number of classifiers which give the positive, negative and neutral sentiment label to each tweet and select the sentiment label which have the highest number of votes.
For each new tweet from the test set, we convert it to 2-dim matrix, if the tweet is longer than  \textit{maxl}, it will be truncated. We then feed it into the ten networks and  pass the results to the voting system.

\textbf{Official ranking:} Our system is ranked fourth over 38 systems in terms of macro-average recall. Table 4 shows the results of our system on the test set of 2016 and 2017.

 \begin{table}[th!]
\centering
\begin{tabular}{|c|c|c|c|}
\hline
Test Dataset& Avg. Recall & Accuracy& F-score\\
\hline
Test 2017 & 0.674&0.652 & 0.665\\
Test 2016 & 0.692& 0.650&0.643\\
\hline
\end{tabular}
\caption{Table 1: Senti17 results on the test sets of 2016 and 2017. }
\end{table}

\section{Conclusion}

We presented our deep learning approach to Twitter sentiment analysis. We used ten convolutional neural network voters to get the polarity of a tweet, each voter has been trained on the same training data using the same word embeddings but different initial weights. The results  demonstrate that our system  is competitive  as it is ranked forth in SemEval-2017 task 4-A.
\bibliography{naaclhlt2016}
\bibliographystyle{naaclhlt2016}

\end{document}